\documentclass[conference]{IEEEtran}
\IEEEoverridecommandlockouts
\usepackage{cite}
\usepackage{amsmath,amssymb,amsfonts}
\usepackage{algorithmic}
\usepackage{graphicx}
\usepackage{textcomp}
\usepackage{xcolor}
\usepackage{url}
\def\BibTeX{{\rm B\kern-.05em{\sc i\kern-.025em b}\kern-.08em
    T\kern-.1667em\lower.7ex\hbox{E}\kern-.125emX}}
\begin{document}

\title{Into the Unknown:\\Self-Learning Large Language Models
}

\author{
% \IEEEauthorblockN{1\textsuperscript{st} Given Name Surname}
% \and
% \and
\IEEEauthorblockN{Teddy Ferdinan}
\IEEEauthorblockA{\textit{Department of Artificial Intelligence} \\
\textit{Wroclaw Tech}\\
Wrocław, Poland \\
teddy.ferdinan@pwr.edu.pl\\
ORCID: 0000-0003-3701-3502
}
\and
\IEEEauthorblockN{Jan Kocoń}
\IEEEauthorblockA{\textit{Department of Artificial Intelligence} \\
\textit{Wroclaw Tech}\\
Wrocław, Poland \\
jan.kocon@pwr.edu.pl\\
ORCID: 0000-0002-7665-6896
}
\and
\IEEEauthorblockN{Przemysław Kazienko}
\IEEEauthorblockA{\textit{Department of Artificial Intelligence} \\
\textit{Wroclaw Tech}\\
Wrocław, Poland \\
kazienko@pwr.edu.pl\\
ORCID: 0000-0001-5868-356X
}
}

\maketitle

\begin{abstract}
We address the main problem of self-learning LLM: the question of what to learn. 
We propose a self-learning LLM framework that enables an LLM to independently learn previously unknown knowledge through self-assessment of their own hallucinations. 
% Hallucination in LLM is a serious problem that hinders public trust in AI. It is often caused by an LLM's lack of knowledge on a given topic. 
% Yet, updating an LLM's knowledge is often very costly and inefficient due to the difficulty of determining exactly what the model already knows and what it does not know yet. 
We introduce a concept called Point in the Unknown (PiU) to identify atomic knowledge unknown to a model, along with four methods for automatic PiUs identification, facilitating the creation of a self-learning loop that focuses exclusively on the absorption of currently unknown knowledge into the model. 
% Using the hallucination score, we introduce a new concept of Points in the Unknown (PiUs), along with one extrinsic and three intrinsic methods for automatic PiUs identification. It facilitates the creation of a self-learning loop that focuses exclusively on the knowledge gap in Points in the Unknown, resulting in a reduced hallucination score. 
% Furthermore, we also present the Self-Learning LLM,
% a concept / a task / a framework ?
% a framework that enables an LLM to learn independently by evaluating its own hallucination, along with the evaluation metrics for gauging an LLM's Self-Learning capability. 
Additionally, we developed evaluation metrics to gauge an LLM's self-learning capability. 
% We also developed evaluation metrics for gauging an LLM's self-learning capability. 
Our experiments revealed that LLMs with at least ~3B parameters that have undergone some instruction training would be able to perform self-learning well. 
% Our experiments revealed that 7B-Mistral models that have undergone instruction finetuning or alignment and RWKV5-Eagle are capable of self-learning considerably well. 
We further proved the effectiveness of self-learning by comparing the performance of a model that has undergone self-learning to a model that has not. 
Our self-learning concept allows more efficient LLM updates and opens new perspectives for LLM knowledge exchange.
% consumption of computing resources for updating LLMs, 
% and in the long-term, open various technological possibilities.
\end{abstract}

\begin{IEEEkeywords}
self-learning, hallucination, LLM, NLP.
\end{IEEEkeywords}

\begin{figure}[!ht]
  \centering
  \fbox{\includegraphics[width=0.465\textwidth]{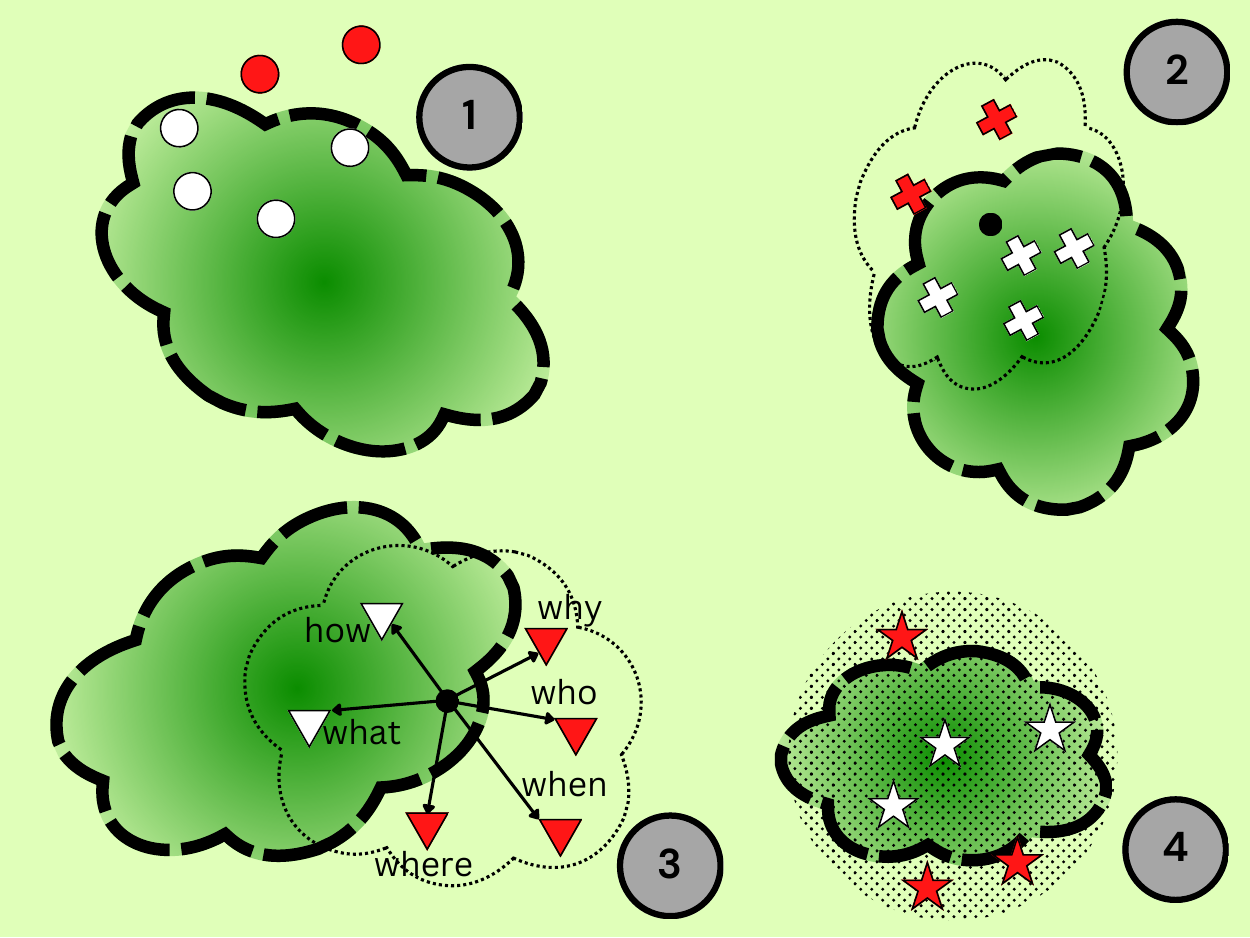}}
  \caption{The illustrative space of knowledge embeddings reduced to two dimensions. It visualizes our four methods for the identification of Points in the Unknown (PiUs), later exploited in the self-learning loop. Dashed lines are the borders of the Known regions (darker green) -- hallucination score thresholds. Out of them are the Unknown regions (lighter green). White points indicate prompts related to knowledge already known to the model, while red points indicate PiUs. Different shapes depict different methods: (1) circles represent extrinsic (external) triggers, i.e., user queries or trending topics; (2) crosses denote open questions-prompts generated by the model itself within a given topic represented by a dotted line; (3) triangles represent the induced questions generated within a topic using 5W+1H; (4) stars indicate the random sampling by selecting random points in the embedding space.}
  \label{fig_methods}
\end{figure}

\begin{figure}[htb]
  \centering
  \fbox{\includegraphics[width=0.465\textwidth]{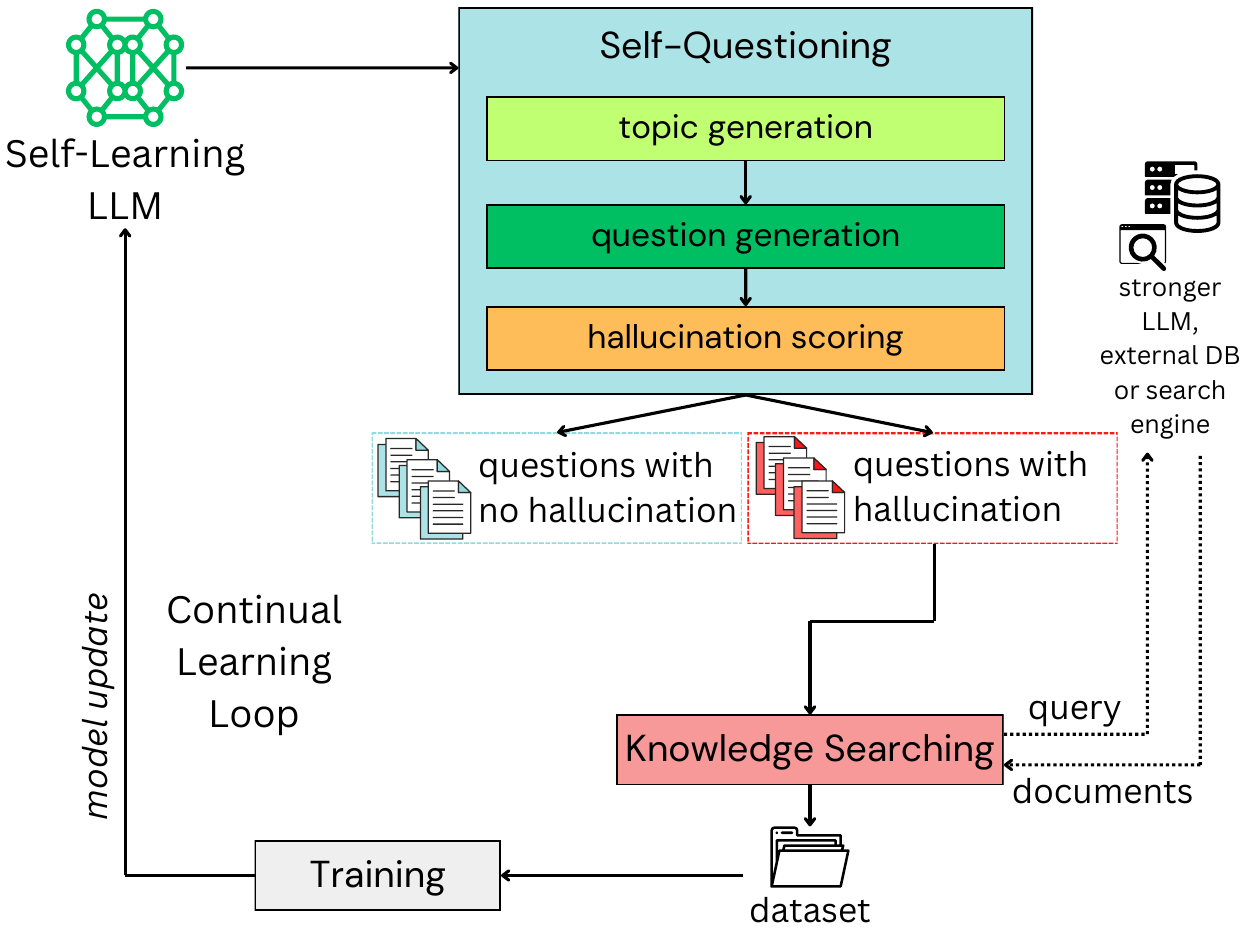}}
  \caption{Illustration of self-learning LLM with intrinsic inspiration.}
  \label{fig_inspiration_intrinsic}

  \bigskip

  \fbox{\includegraphics[width=0.465\textwidth]{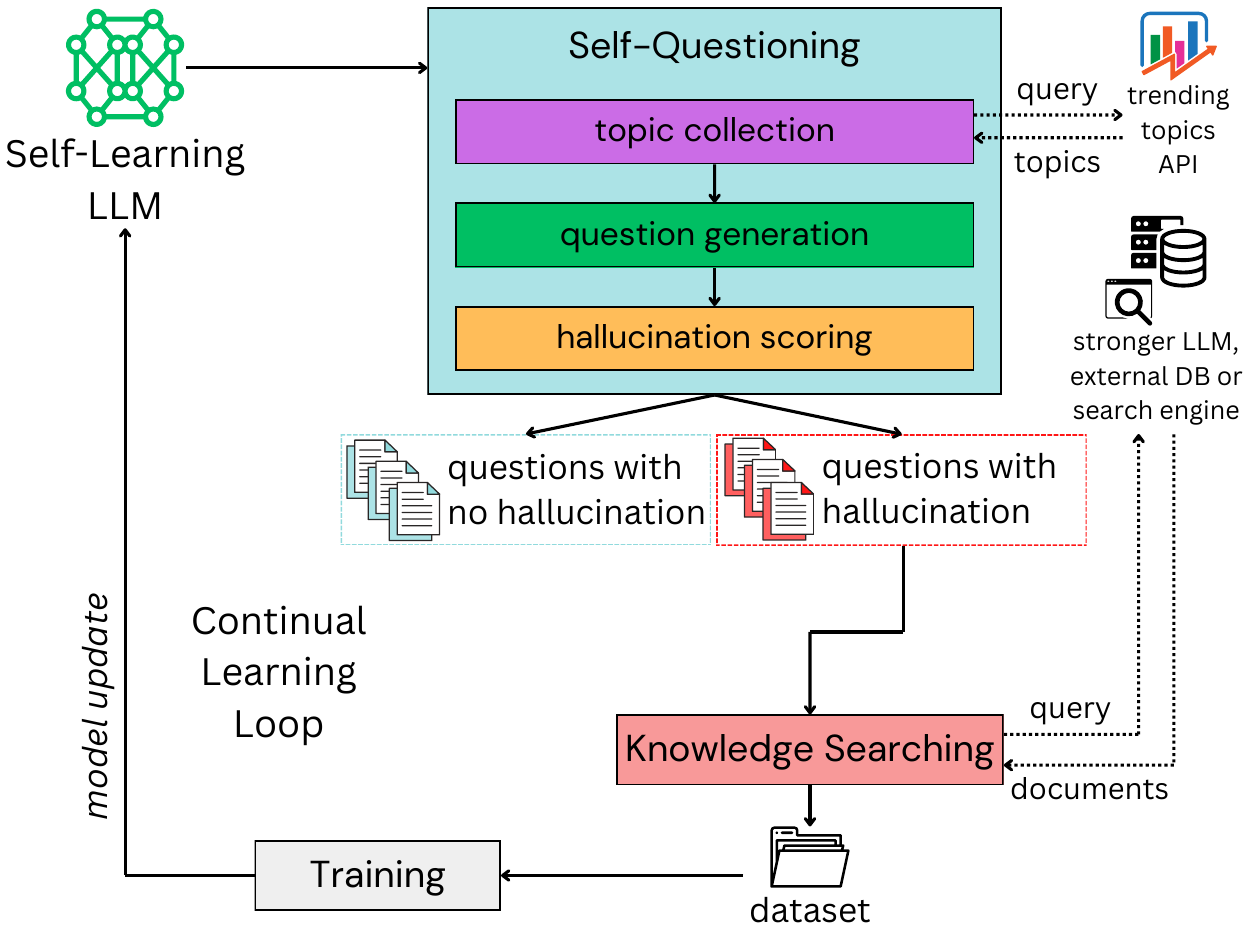}}
  \caption{Illustration of self-learning LLM with extrinsic inspiration.}
  \label{fig_inspiration_extrinsic}
\end{figure}

\section{Introduction}
Commonly, Large Language Models (LLMs) are pre-trained on large textual corpora and then fine-tuned using additional data to be better adjusted to a given policy or domain. Simultaneously, other methods have been developed, which are based on additional knowledge provided to the model directly in the prompt, such as Retrieval Augmented Generation (RAG). In this paper, we explore a different concept: self-learning LLM. It is the persistent acquisition of new knowledge by the model without data provision, taking advantage of three fundamental mechanisms that are integrated in a continuous loop: (1) identification of what knowledge to learn, (2) gathering new relevant data, and (3) continuous model training.

The novelty of this self-learning framework and what sets it apart from traditional continual learning is the ability of the system to determine gaps in its own knowledge and construct the dataset on which it can train itself without repeating knowledge that it already knows. The self-learning LLM utilizes hallucination on simple questions as an indicator of unknown knowledge; while hallucination can be caused by a wide range of factors, one of the main reasons for hallucination on simple questions is due to the model not possessing the factual knowledge that can answer the question~\cite{onoe2022_unseen, ji2024_unseen}. For example, if a model hallucinates when asked, "Who won the gold medal from men's singles badminton in the 2024 Summer Olympics?" it is likely that the model's training did not include the information related to the event.
% Notably, while a RAG-based system would be able to circumvent the issue of incomplete or outdated training data, RAG does not actually update the model's knowledge, eventually limiting the model's applicability.

The self-learning LLM is particularly useful for having an up-to-date model without training a new one from scratch while also minimizing human involvement, greatly reducing development costs. It can be applied in application domains where new facts continuously come up. One example is personalized sentiment prediction and emotion recognition \cite{milkowski2023modeling,koptyra2023clarin,kocon2023chatgpt,kanclerz2023pals,kazienko2023human,kocon2023deep,ferdinan2024fortifying,fan2024fusing}. When new users enter the system, the LLM would be able to perform self-learning only on data related to the new users. Another potential application is the knowledge exchange between LLMs, allowing the knowledge of one LLM to be learned by another LLM automatically.

% Methods for the Unknown detection: \\
% 1. External Prompt (extrinsic) \\
% 2. Open Generation (intrinsic) \\
% 3. Induced Generation (intrinsic) \\
% 4. Oracle-selected (intrinsic) \\

The contribution presented in this paper covers: (1) the concept of Points in the Unknown (PiUs) to identify knowledge unknown to a model; (2) four different methods to identify PiUs; (3) metrics to gauge the capability of a model to conduct self-learning; (4) design of the self-learning LLM; (5) experimental validation of the methods to identify PiUs; (6) experimental validation of the self-learning LLM; (7) software and data for reproducibility.

\section{Related Work}
\label{sec:related_work}
Hallucination in the context of LLM is the problem of nonsensical or unfaithful information produced by a generative model. Some works have studied the causes of hallucination~\cite{kandpal2023_longtail, lee2022_deduplicating, onoe2022_unseen, ji2024_unseen}, as well as the detection methods~\cite{lin2022_truthfulqa, min2023_factscore, azaria2023_lying, manakul2023_selfcheckgpt, cao2023_autohall, yin2023_selfaware}. Some solutions for overcoming hallucination are proposed in~\cite{ji2023_selfreflection, luo2023_selffamiliarity, li2023_inferencetime, ji2023_rho, liu2023_roe, liang2024_dreamcatcher, tian2024_factuality, chuang2024_dola, chuang2024_lookbacklens}. Meanwhile, in Retrieval Augmented Generation (RAG)~\cite{lewis2020_RAG}, hallucination is avoided by supplying the prompt with some context retrieved from an outside source, allowing more factual generation without updating the model.

Continual Learning is a training paradigm where the model is subjected to various tasks sequentially; in the context of LLM, the tasks typically comprise domain-specific datasets~\cite{jang2022_continualLM, jin2022_continualLM, ke2023_continualLM}. One challenge is preventing catastrophic forgetting, in which the model loses knowledge from previous tasks~\cite{mccloskey1989_catastrophic, ratcliff1990_catastrophic}. Solutions for adding or editing knowledge while avoiding catastrophic forgetting have been proposed~\cite{kirkpatrick2017_ewc, zhu2020_memories, sinitsin2020_editable, ke2021_ctr, mitchell2022_serac}.

\section{Why Self-Learning is Needed}
Hallucination is a serious problem that hinders many LLM applications. One of its main reasons is the model's lack of knowledge on a given topic, or the model's knowledge on the given topic has become obsolete~\cite{onoe2022_unseen, ji2024_unseen}. This problem is typically overcome either by providing the knowledge as additional context~\cite{lewis2020_RAG} without subjecting the model to more learning or by continuous training using new data. With the former, the model would still become outdated after some period of time because most of its knowledge would have become obsolete. With the latter, there is a problem of determining what the model already knows and what it does not know yet, especially if there is limited information on the model's past training data; if the training focuses on knowledge already acquired by the model, it does not solve the hallucination problem, and we needlessly waste a lot of computing resources by merely repeating known knowledge.

Therefore, it is essential to identify the knowledge known and unknown to the model in order to conduct continuous training as efficiently as possible. The self-learning LLM would be able to distinguish unknown knowledge to create a dataset for its own training automatically. This would reduce the required computing resources as well as human involvement, making continuous training much more efficient. One major concern would be catastrophic forgetting; however, in Section~\ref{sec:experiment2}, we will show that catastrophic forgetting can be avoided by carefully choosing the training technique and architecture.

\section{The concept of Point in the Unknown (PiU)}
We introduce the concepts of \textit{The Known} and \textit{The Unknown} to define the problem more precisely. First of all, if we represent each atomic piece of human knowledge as a vector, it would be possible to form an abstract space that includes all pieces of human knowledge. We call this abstract space as \textit{Human Knowledge Space} (HKS).
% By using vectorial representation, we can form an abstract space in which each point represents an atomic piece of knowledge. 
% Human knowledge is an abstract space formed by vectorial knowledge representations; each point in the space represents an atomic piece of knowledge.

\textit{\textbf{The Known}} refers to an area in \textit{HKS} where our LLM does not hallucinate, i.e., it possesses sufficient knowledge related to this region. We call each point in such an area as \textit{Point in the Known} (PiK, plural: PiKs).
% A PiK represents a knowledge that our LLM already possesses.

\textit{\textbf{The Unknown}} refers to an area in \textit{HKS} where our LLM hallucinates; each point in it is called \textit{Point in the Unknown} (PiU, plural: PiUs). A PiU represents an atomic piece of knowledge that our LLM lacks, which we want the model to identify and acquire.

Finding the boundaries between \textit{The Known} and \textit{The Unknown} is non-trivial. However, if we utilize hallucination on simple questions as an indicator of unknown knowledge, it would be possible to adopt a hallucination detection method that provides a numerical scoring system to approximate such boundaries. One such method is 
SelfCheckGPT~\cite{manakul2023_selfcheckgpt}, a sampling-based hallucination detection method. Given a prompt $x$, it first makes the LLM generate the \emph{main passage}, a response to the prompt generated using greedy decoding. Then, it makes the LLM generate $n$ samples; these are responses to the prompt generated using multinomial sampling with the temperature set to 1.0. Next, SelfCheckGPT checks the consistency between the main passage and the samples to output a hallucination score $h(x) \in [0,1]$, where $h(x)=1$ indicates certain hallucination (lack of any knowledge about $x$) and $h(x)=0$ means no hallucination (certain knowledge of $x$). In this case, the median between the two values, $\mathit{LIMIT}=0.5$, would serve as an intuitive and reasonable constant threshold to approximate the boundary between \textit{The Known} and \textit{The Unknown}. Therefore:

\begin{align*}
\mathit{Known} &= \{x : h(x) < \mathit{LIMIT}; x \in \mathit{HKS}\} \\
Unknown &= \{x : h(x) \geq \mathit{LIMIT}; x \in \mathit{HKS}\}
\end{align*}

As mentioned before, $\mathit{PiK} \in \mathit{Known}$ and $\mathit{PiU} \in \mathit{Unknown}$. Note that $\mathit{Known} \cap \mathit{Unknown} = \emptyset$ and $\mathit{Known} \cup \mathit{Unknown} = \mathit{HKS}$, even though the sizes of $\mathit{Known}$ and $\mathit{Unknown}$ may change after training.

Alternatively, we could have exploited some other methods for $h(x)$, like SAPLMA~\cite{azaria2023_lying}, DoLa~\cite{chuang2024_dola}, or Lookback Lens~\cite{chuang2024_lookbacklens}. However, SAPLMA and DoLa would require access to the model's internal features, limiting the feasibility, while Lookback Lens is primarily aimed at RAG applications. Therefore, we decided to stick with SelfCheckGPT, at least for the current work.

SelfCheckGPT has several variants. While the LLM-based variant using GPT-3.5-turbo gave the best results in the original paper, we chose the NLI-based variant. It is recommended if dependency on another LLM is not desired and works faster while still providing decent performance. This variant works by treating each sentence in the main passage as a hypothesis and each sample as a potential premise, using probabilities of entailment and contradiction to output a normalized score bounded between 0.0 and 1.0. In our experiments, we generate 10 samples per main passage, as the original paper's ablation study showed that performance plateaus at 10 samples, with no significant gain from using more.
% SelfCheckGPT comes in several variants. While the LLM-based variant using gpt-3.5-turbo gave the best results in the original paper, we instead chose the NLI-based variant. It has been recommended as an alternative if dependency on another LLM is not desired. It also works considerably faster while still providing decent performance. The NLI-based variant of SelfCheckGPT works by taking each sentence in the main passage as a hypothesis and checks each sample as the premise to it, considering only the probabilities of entailment and contradiction to output a normalized score bounded between 0.0 and 1.0. In our experiments, we always generate 10 samples to check each main passage; the ablation study in the original paper showed that the performance of the NLI-based variant of SelfCheckGPT would already reach a plateau when using 10 samples, and there is insignificant gain from using more samples.

\section{Methods for Identification of PiUs}
%Methods for Identification of Points in the Unknown (Methods for IPU) (MIPU)
Identification of PiUs can be done by evaluating the model's hallucination score $h(x)$ on some question-prompt $x$. The question, or the inspiration for the question, can come from outside the system, supplied by an external entity; such a method would be considered to have an \textit{extrinsic} nature. On the other hand, we can create a built-in \emph{oracle} inside the system that guides automatic question generation in a bicameral-mind manner~\cite{jaynes1990_bicameralmind}, in which case the method is deemed \textit{intrinsic}. We propose one extrinsic and three intrinsic methods for PiU identification, Figure~\ref{fig_methods}. All of them differently generate $x$, which, if tested according to $h(x) \geq LIMIT$, may be identified as PiU.

\subsection{External Prompt (extrinsic)}
There are some existing ideas related to collecting prompts that cause hallucination and constructing a dataset based on them for finetuning~\cite{zhu2020_memories, cao2023_autohall, tian2024_factuality, liang2024_dreamcatcher}. However, they require manual curation of the prompts collected from users or datasets, so the model does not learn fully independently.

In our approach, we utilize an external API to collect trending topics as inspiration for formulating concrete questions. Every item returned by the API is a list of related phrases; we treat each list as a single topic. Then, the model is asked to generate a specific question $x$ based on each topic. Next, the model is asked to answer the question in order to evaluate its hallucination $h(x)$. 
% Notably, an oracle is still used in the system to control question generation and hallucination scoring, yet the topics that trigger the questions come from outside the system. 
This approach allows the model to continuously learn by itself as long as the external API is available; however, the tested space is limited only to trending topics indirectly provided by humans.

\subsection{Open Generation (intrinsic)}
In this method, the oracle asks the model to propose some topics to learn about. Then, the oracle asks the model to consider those topics and formulate one question ($x$), to which the model believes it does not know the answer. Finally, the oracle asks the model to answer the question $x$ in order to evaluate its hallucination $h(x)$.

\subsection{Induced Generation (intrinsic)}
It is based on \emph{Five Ws} and \emph{How}, which are widely considered basic questions for information comprehension and data gathering. Here, the oracle also asks the model to propose some topics. Then, the oracle asks the model to formulate a question $x$ using a particular question word; this is repeated six times for \emph{what}, \emph{who}, \emph{why}, \emph{where}, \emph{when}, and \emph{how}, resulting in six different questions. Finally, the oracle also asks the model to answer each question and evaluate its hallucination $h(x)$.

\subsection{Oracle-Selected (intrinsic)}
This method starts by constructing a topic embedding space, which contains all candidate topics represented in a vectorial form. Then, the oracle randomly selects a point in the topic embedding space and samples the nearest neighbors to that point. This results in a set of oracle-selected topics. Next, the oracle asks the model to consider those topics and formulate one question $x$. Afterward, the oracle asks the model to answer this question and evaluates the hallucination $h(x)$.

\section{Self-Learning LLM}
Self-learning is a process where our LLM identifies its own PiUs, searches for the knowledge related to these PiUs, and trains itself on the collected data. It is made possible by incorporating three fundamental mechanisms in a continuous loop: Self-Questioning, Knowledge Searching, and Model Training. Self-learning LLM with an intrinsic method is illustrated in Figure~\ref{fig_inspiration_intrinsic}, while self-learning LLM with an extrinsic method is illustrated in Figure~\ref{fig_inspiration_extrinsic}.

% Self-Learning has some similarities to traditional continual learning. The key difference lies in the LLM system, which asks questions by itself and evaluates whether it knows the answer or not. Another difference is the dynamic and integral dataset construction process in Self-Learning; in traditional continual learning, the dataset construction process is typically out of the scope. Nevertheless, training techniques and model architectures from traditional continual learning can possibly work in Self-Learning to allow more efficient training and avoid catastrophic forgetting, one of which will be presented in Section~\ref{sec:experiment2}.

\subsection{Self-Questioning}
Self-Questioning is generally performed through topic generation (or topic collection), question generation, and hallucination scoring. Depending on the selected method for the identification of PiUs, the logical implementation of Self-Questioning may differ slightly. Self-Questioning is repeated in a loop for $N$ iterations; the primary output consists of a list of generated questions with hallucination ($Q_H$) and a list of generated questions with no hallucination ($Q_{NH}$). In other words, $Q_H \subset Unknown$ and $Q_{NH} \subset Known$. A question is categorized into $Q_H$ if the hallucination score of the main passage is greater than 0.5 or into $Q_{NH}$ otherwise.
% Appendix~\ref{sec:appendix_methods} provides illustrations of such logical implementations with External Prompt, Open Generation, Induced Generation, and Oracle-Selected.

% \subsection{Meaningfulness Filter}
% An optional component before Knowledge Searching is the Meaningfulness Filter. It could be a separately trained model, whose goal is: (1) to remove obviously meaningless PiUs like \textit{"How fast cats fly on Mars?"}; (2) to implement goals and policy of self-learning, e.g., consider only knowledge (PiUs) relevant to a given domain; (3) to overcome limitations of hallucination score, e.g., when the LLM's response is \textit{"I do not know,"} and its $h(x) \approx 0$. In our experiments, we did not implement the Meaningfulness Filter because we considered any question in any domain as valid.

\subsection{Knowledge Searching}
After Self-Questioning and filtering, Knowledge Searching queries an external source to collect knowledge that can answer $Q_H$ in order to build the dataset $D_{train}$. Notably, Knowledge Searching may be implemented inside Self-Questioning by immediately searching for the answer to an individual question whenever a hallucination is detected. However, having Knowledge Searching separately is more practical, allowing additional processing without creating a bottleneck for the Self-Questioning process.

% However, there is an optional component before Knowledge Searching, which is the Meaningfulness Filter. In domain-specific applications, some questions generated by the model may be considered meaningless or not relevant. The Meaningfulness Filter is a classifier trained to distinguish meaningful questions for the specific application domain, so that meaningless questions will not be included for Knowledge Searching. We did not implement the Meaningfulness Filter in our experiments because we considered any question in any domain as valid.

\subsection{Model Training}
The model is trained on $D_{train}$ to absorb the knowledge for answering $Q_H$. Once training is done, PiUs should become PiKs, effectively increasing the Known regions and decreasing the Unknown regions of the model. Afterward, the next self-learning cycle can start to find more PiUs.

\section{Metrics for choosing a base model for self-learning}
Self-learning requires a pretrained model that already possesses a sufficient understanding of instructions. In our initial experiments, we observed that some models are better at asking the "right" questions (questions on which the model would actually hallucinate) than others. The ability of a model to ask such questions would directly affect the success of self-learning. Therefore, we propose some metrics to evaluate the capability of a model to self-learn.

\subsection{Curiosity Score}
It measures how likely a model would explore different questions. A high Curiosity score indicates the model tends to ask unique, different questions over multiple iterations of Self-Questioning and, hence, is more likely to explore Unknown regions. It is calculated as follows:
\begin{equation}
Cur = \frac{\#Q_{unique}}{\#Q}
\end{equation}
\noindent where $\#Q_{unique}$ is the number of unique questions generated, and $\#Q$ is the total number of questions generated.

$\#Q_{unique}$ is determined by doing HDBSCAN clustering~\cite{mcinnes2017_hdbscan} with the "leaf" cluster selection method on the question embeddings, counting the number of clusters and outliers. The question embeddings are acquired by using all-MiniLM-L12-v2 from Sentence Transformers~\cite{reimers2019_senttrans} on the questions generated by our self-learning LLM.

\subsection{Knowledge-Limit Awareness Score}
It indicates how likely a model would come up with a question that it cannot answer without hallucination during Self-Questioning -- how likely a model is aware of its own knowledge limitation. It is calculated as follows:
\begin{equation}
Kaw = \frac{\#Q_{H}}{\#Q}
\end{equation}
\noindent where $\#Q_{H}$ is the number of generated questions with hallucination, and $\#Q$ is the total number of questions generated.

\subsection{Brevity Coefficient}
It is used to penalize the evaluation when the brevity constraint is violated (e.g. when the model fails to formulate one concrete question without elaboration). The brevity coefficient $brev$ is calculated as follows:

\begin{equation}
  ideal\_len = 100
\end{equation}
\begin{equation}
  \Delta_{len} = \frac{\sum_{i=1}^{n_{text}} |len_i - ideal\_len|}{n_{text}} \\
\end{equation}
\begin{equation}
  brev = 
  \begin{cases}
    0,& \text{if }  \Delta_{len} \geq 100\\
    1,& \text{if } 0 \leq \Delta_{len} \leq 50\\
    1 - \frac{\Delta_{len}}{ideal\_len} + \frac{1}{2},& \text{otherwise}
  \end{cases}
\end{equation}

\noindent where $ideal\_len$ is the assumed ideal average text length measured in character count, $len_i$ is the length of the $i$~-th text, $n_{text}$ is the number of texts, and $\Delta_{len}$ is the average difference between $ideal\_len$ and the text lengths.

The brevity coefficient has been designed to decrease gradually in a linear manner as the average text length goes further from the range [50,150] before dropping immediately to zero when the average text length becomes too far from the ideal. The thresholds of 50 and 150 are roughly based on the research by Miller, Newman, and Friedman~\cite{miller1958_textlen}; such thresholds represent the approximated minimum and maximum lengths of a typical sentence in the English language. We also found in our initial exploration that these thresholds are suitable for the task of generating a single concrete question in the English language.

\subsection{Self-Learning Capability (SLC) Score}
It is a simple average of the two components, Curiosity score and Knowledge-Limit Awareness score, multiplied by the brevity coefficient. Such an aggregation is meant to allow easy comparison between models. It is calculated as follows:
\begin{equation}
SLC = brev * \frac{Cur + Kaw}{2}
\end{equation}
\noindent where $brev$ is the brevity coefficient, $Cur$ is the Curiosity score, and $Kaw$ is the Knowledge-Limit Awareness score.

A higher SLC score indicates the model is more suitable for self-learning, being more likely to ask different questions and also to ask questions on which it would hallucinate. Note, however, that a model that has been trained extensively to retain a huge amount of knowledge may struggle to ask questions on which it would actually hallucinate. In such a case, the model may achieve a relatively low SLC score, which means "plain" self-learning would not be beneficial for the model. This brings the question about an entirely new learning task: how to make a model aware of its own knowledge limitation so that it would ask questions that would expand its own knowledge; it will be discussed further in Section~\ref{sec:discuss_applications}.

\section{Experimental Setup}
All experiments were conducted with Python 3.10. The machine featured 8 CPU cores, 300GB RAM, and one NVIDIA H100 94GB. 
The full code and archived results are available publicly with GPL-3.0 license\footnote{\url{https://github.com/teddy-f-47/self-learning-llm-public}}.
% The full code and archived results are available publicly with GPL-3.0 license\footnote{\textcolor{brown}{blind policy -- Anonymous GitHub: \url{https://anonymous.4open.science/r/self-learning-llm-public-5BAE}}}.
All work is intended only for scientific research.

\section{Experiment 1: Self-Learning Capability}
\label{sec:experiment1}
% \section{Experiments}
Experiments were performed to investigate the feasibility of creating a self-learning LLM using different pretrained models. In addition, we would also like to investigate the effectiveness of different methods for the identification of PiUs.

\subsection{Models}
The details of seven pretrained models, some of which have also been instruction-finetuned or aligned, are presented in Table~\ref{table_model_detail}. The column "HF Name" in the table provides the models' names on the HuggingFace platform. Mistral-dpo is a 7B-Mistral model that has been aligned with DPO~\cite{rafailov2023_dpo} by Intel. Mistral-instruct is a 7B-Mistral model that has been instruction-finetuned by Mistral. Both of them are actually based on the same pretrained model, which is codenamed mistral-base~\cite{jiang2023_mistralbase} in our experiments. We also included TinyLlama~\cite{zhang2024_tinyllama}, Phi-3-small, Phi-3-mini~\cite{microsoft2024_phi3}, and RWKV5-Eagle~\cite{peng2024_rwkv5} for comparison. Finally, we defined a baseline, which is a deterministic dummy model which would always respond with the same text $f(x)$ when given the same prompt $x$, i.e., $f(x_1) = f(x_2) \iff x_1 = x_2$.

\begin{table*}[hbt]
  \centering
  \caption{Details of LLMs used in the experiments. "Model Name" shows the model's codename in our work, while "HF Name" provides the model's ID available on the HuggingFace platform.}
  \label{table_model_detail}
  \begin{tabular}{|l|l|r|l|}
    \hline
      \textbf{Model Name} & \textbf{HF Name} & \textbf{Number of Parameters} & \textbf{Finetuned or Aligned?} \\ \hline
      mistral-dpo & Intel/neural-chat-7b-v3-3 & 7.2B & Yes - DPO \\ \hline
      mistral-instruct & mistralai/Mistral-7B-Instruct-v0.2 & 7.2B & Yes - SFT \\ \hline
      mistral-base & mistralai/Mistral-7B-v0.1 & 7.2B & No \\ \hline
      rwkv5-eagle & RWKV/v5-Eagle-7B-HF & 7.5B & Partial instruct tuning \\ \hline
      phi-3-small & microsoft/Phi-3-small-8k-instruct & 7.4B & Yes - SFT and DPO \\ \hline
      phi-3-mini & microsoft/Phi-3-mini-4k-instruct & 3.8B & Yes - SFT and DPO \\ \hline
      tiny-llama-chat & TinyLlama/TinyLlama-1.1B-Chat-v1.0 & 1.1B & Yes - SFT and DPO \\ \hline
    \end{tabular}
\end{table*}

\subsection{Data}
The experiment with the Open Generation method involved 3000 self-questioning iterations, resulting in 3000 total generated questions. The same was true for the experiment with the Oracle-Selected method. Meanwhile, the experiment with Induced Generation involved 500 self-questioning iterations to produce 3000 total generated questions. The experiment with External Prompt utilized the Google Trends API from SerpApi\footnote{\url{https://serpapi.com/google-trends-api}} and involved 10 self-questioning iterations; since the list of items returned by the API from each request had a variable length, this resulted in 576 generated questions. To allow fair comparison between models, we cached the received trending topics so that all models were given the same topics for self-questioning. All data is in the English language.

\subsection{Results}
Table \ref{table_exp_result} enumerates the experiment results with different methods. "Cur" is the Curiosity score, "Kaw" is the Knowledge-Limit Awareness score, "brev" is the brevity coefficient, and "SLC" is the Self-Learning Capability score.

\begin{table}[hbt]
  \centering
  \caption{Experimental results. Each row presents each model's evaluation result using a particular method for identification of PiUs. "Cur" indicates the Curiosity score, "Kaw" indicates the Knowledge-Limit Awareness score, "brev" indicates the brevity penalty, and "SLC" is the Self-Learning Capability score.}
  \label{table_exp_result}
  \begin{tabular}{|l|l|r|r|r|r|r|}
    \hline
      \textbf{Model Name} & \textbf{Method} & \textbf{Cur} & \textbf{Kaw} & \textbf{brev} & \textbf{SLC} \\ \hline
      mistral-dpo & Open Gen. & 0.73 & 0.04 & 1.00 & 0.38 \\
      mistral-dpo & Induced Gen. & 0.75 & 0.08 & 1.00 & 0.42 \\
      mistral-dpo & Oracle-Select. & 0.96 & 0.18 & 1.00 & 0.57 \\
      mistral-dpo & Ext. Prompt & 0.73 & 0.12 & 1.00 & 0.42 \\ \hline
      mistral-instruct & Open Gen. & 0.39 & 0.29 & 0.99 & 0.34 \\
      mistral-instruct & Induced Gen. & 0.63 & 0.17 & 0.59 & 0.24 \\
      mistral-instruct & Oracle-Select. & 0.97 & 0.31 & 0.92 & 0.58 \\
      mistral-instruct & Ext. Prompt & 0.60 & 0.26 & 1.00 & 0.43 \\ \hline
      mistral-base & Open Gen. & 0.82 & 0.81 & 0.00 & 0.00 \\
      mistral-base & Induced Gen. & 0.79 & 0.82 & 0.00 & 0.00 \\
      mistral-base & Oracle-Select. & 0.95 & 0.76 & 0.00 & 0.00 \\
      mistral-base & Ext. Prompt & 0.95 & 0.79 & 0.00 & 0.00 \\ \hline
      rwkv5-eagle & Open Gen. & 0.90 & 0.41 & 0.59 & 0.39 \\
      rwkv5-eagle & Induced Gen. & 0.92 & 0.48 & 0.75 & 0.53 \\
      rwkv5-eagle & Oracle-Select. & 0.97 & 0.45 & 0.93 & 0.66 \\
      rwkv5-eagle & Ext. Prompt & 0.70 & 0.45 & 1.00 & 0.58 \\ \hline
      phi-3-small & Open Gen. & 0.47 & 0.09 & 1.00 & 0.28 \\
      phi-3-small & Induced Gen. & 0.59 & 0.21 & 1.00 & 0.40 \\
      phi-3-small & Oracle-Select. & 0.94 & 0.33 & 1.00 & 0.63 \\
      phi-3-small & Ext. Prompt & 0.76 & 0.38 & 1.00 & 0.57 \\ \hline
      phi-3-mini & Open Gen. & 0.72 & 0.08 & 1.00 & 0.40 \\
      phi-3-mini & Induced Gen. & 0.76 & 0.21 & 1.00 & 0.49 \\
      phi-3-mini & Oracle-Select. & 0.96 & 0.36 & 1.00 & 0.66 \\
      phi-3-mini & Ext. Prompt & 0.68 & 0.47 & 1.00 & 0.57 \\ \hline
      tiny-llama-chat & Open Gen. & 0.92 & 0.22 & 0.00 & 0.00 \\
      tiny-llama-chat & Induced Gen. & 0.88 & 0.20 & 0.00 & 0.00 \\
      tiny-llama-chat & Oracle-Select. & 0.99 & 0.29 & 0.00 & 0.00 \\
      tiny-llama-chat & Ext. Prompt & 0.81 & 0.34 & 0.00 & 0.00 \\ \hline
      baseline & Open Gen. & 0.0003 & 0.00 & 0.00 & 0.00 \\
      baseline & Induced Gen. & 0.002 & 0.00 & 0.00 & 0.00 \\
      baseline & Oracle-Select. & 0.98 & 0.00 & 0.00 & 0.00 \\
      baseline & Ext. Prompt & 0.62 & 0.00 & 0.00 & 0.00 \\ \hline
    \end{tabular}
\end{table}

\textbf{Instruction training}. Our experimental results suggest that instruction training, either inflicted through supervised finetuning (SFT) or some alignment technique, plays a crucial aspect in allowing self-learning. Instruction training enables the model to understand the self-learning instruction to form a concise question. As shown by mistral-base's results, the non-finetuned model would always fail to formulate a concise question, preventing an effective self-learning to take place. On the other hand, finetuning also reduces the tendency of a model to hallucinate: the Knowledge-Limit Awareness scores of mistral-dpo and mistral-instruct were always lower than mistral-base's. This is because the mistral-base does not yet know how to answer a lot of questions properly. Self-learning could be beneficial for such a model if it was able to formulate concise questions. Interestingly, rwkv5-eagle, which has not been finetuned, consistently achieved high SLC scores; this can be attributed to the nature of its pretraining data, which contained some instruction examples, allowing the model to understand the command to form concise questions.

\textbf{Model size}. The model size is also quite important; if the model is too small, it may lack the capacity to understand and follow instructions properly. This is indicated by the results from tiny-llama-chat. Although it has undergone instruction training, it still often fails to generate concise questions without excessive elaboration. On the other hand, the phi-3-mini is slightly larger, and it was able to formulate concise questions for self-learning.

\textbf{Intrinsic and Extrinsic Inspiration}. In a real-world scenario, choosing the kind of method for the identification of PiUs primarily depends on the use-case requirements and constraints. For instance, if keeping the model updated with the latest popular news is pivotal, then an extrinsic method would be best. Conversely, if dependency on an external entity is not desired, or if finding all of the model's PiUs is more important, then an intrinsic method is arguably better. In terms of the effectiveness of different methods, we can find some interesting findings from the experiments.

Open Generation and Induced Generation are generally less effective compared to the other two methods because they rely on the topics proposed by the model itself. Depending on the model's past training data, some topics may have a very high probability of being generated, while others are very low. However, it might be possible to make these methods more effective by increasing the temperature of the multinomial sampling during topic generation. This requires further investigation and is within our future directions. Meanwhile, Oracle-Selected allowed rwkv5-eagle and phi-3-mini to achieve the highest SLC score. Its inherent randomness led to the exploration of a wide range of topics, including some obscure ones, making it particularly effective.

\section{Experiment 2: Model Performance after Self-Learning}
\label{sec:experiment2}
This section provides a simple demonstration of one full self-learning cycle. The goal of the experiments is to compare the performance of a model that has undergone self-learning against its counterpart that has not performed self-learning.

\subsection{Models}
For this experiment, we used mistral-instruct as the base of the self-learning LLM. The training was conducted using LoRA~\cite{hu2022_lora} and the DPO trainer. By using LoRA, we were able to explore two architectural approaches: (1) Plain LoRA and (2) Dynamic-Adapter (Dyn-Adapt). In the case of Plain LoRA, the adapter was simply merged into the base model after training, effectively altering the model's weights. Meanwhile, Dyn-Adapt was inspired by SERAC \cite{mitchell2022_serac} and DAP-Adapter \cite{ke2023_continualLM}; in this case, the adapter was not merged. Each cycle of self-learning would produce a new adapter containing the recently learned knowledge. During inference, Dyn-Adapt would enable the most suitable adapter for a given prompt when necessary or disable all adapters if the base model is deemed best for answering the prompt. The adapter enabling was controlled by a router-classifier model, which was trained on a mix of samples from $Q_H$ and $Q_{NH}$.

\subsection{Data}
We used the output from the Oracle-Selected experiment with mistral-instruct for self-learning. Of the 3,000 total generated questions, 930 were classified into $Q_H$. After deduplication, 922 unique questions were identified and became the basis for creating $D_{train}$ for training. We utilized the model \verb|gpt-4o-2024-05-13|~\cite{openai2024_gpt4} to find the answers to these questions and verified the quality of the answers manually. $D_{train}$ was then constructed as a preference dataset consisting of questions, chosen answers, and rejected answers. We distinguished cases where \textit{the model was unsure} and \textit{the model did not know} by using \verb|gpt-4o-2024-05-13| as a judge to compare the similarity between the model's original predictions and the ground truth. The model was \textit{unsure} when the original prediction was similar to the ground truth despite the high hallucination score; in this case, we used the original prediction as the chosen answer and "-" as the rejected answer. Meanwhile, the model \textit{did not know} when the original prediction was indeed different from the ground truth; in this case, the ground truth was selected as the chosen answer while the model's original prediction was put as the rejected answer. From 922 unique questions, we found 283 \textit{unsure} cases and 639 \textit{did not know} cases.

For evaluation, we used three datasets: $Q_H$, Wiki, and Alpaca. The $Q_H$ dataset is the same 930 questions from the Oracle-Selected experiment and was used for the sake of checking if hallucination on these questions would decrease after self-learning. The Wiki dataset was acquired from the 2023-12-01 dump of Simple English Wikipedia~\cite{wikidump}\footnote{\url{https://huggingface.co/datasets/olm/wikipedia}}, serving as a test dataset to observe catastrophic forgetting; we randomly selected 1,000 rows from the dataset for evaluation. Finally, we also used the cleaned version of Alpaca~\cite{stanford2023_alpaca}\footnote{\url{https://huggingface.co/datasets/yahma/alpaca-cleaned}} as a test dataset to observe if the model would lose unrelated knowledge after self-learning; also 1,000 rows were randomly selected for evaluation.

\subsection{Training Hyperparameters}
The model was trained on $D_{train}$ for three epochs. The learning rate was set to 3e-5. With a micro-batch size of 4 and gradient accumulation of 8, the effective global batch size was 32. Finally, for the LoRA configuration, we used r=64, alpha=128, dropout=0.05, and bias="none", targeting the $q$ and $v$ projection modules.

\subsection{Metrics}
On the $Q_H$ dataset, besides measuring the hallucination score, we also used ROUGE-Lsum \cite{lin2004-rouge} to measure the similarity between the model's answer and the ground truth acquired from Knowledge Searching. Additionally, we used \verb|gpt-4o-2024-05-13| to judge the correctness of the model's answer relative to the ground truth. The LLM-Judge was prompted to answer "yes" if the model's answer could capture the meaning contained in the ground truth; "partly" if the model's answer could be considered partially correct relative to the ground truth; "no" if the model's answer was completely different from the ground truth. The responses from the LLM-Judge were then normalized: "yes" answers were converted to 1.0; "partly" answers were converted to 0.5; "no" answers, as well as any other response not following the instruction, were converted to 0.0.

On the Wiki dataset, we measured the perplexity, a commonly used metric to estimate the language modeling capability of an LLM. A drastic increase in perplexity after self-learning would indicate catastrophic forgetting. Finally, on the Alpaca dataset, we also used ROUGE-Lsum and LLM-Judge.

\subsection{Results}
Table~\ref{table_self_learning_results_mistral} presents the full evaluation results. The column "Baseline" shows the metric values before self-learning.

\begin{table}[hbt]
  \centering
  \caption{Mistral-Instruct evaluation results before and after self-learning. "Baseline" shows the metric values before self-learning; "LoRA" shows the metric values after self-learning using Plain LoRA; "Dyn-Adapt" shows the metric values after self-learning using Dynamic-Adapter.}
  \label{table_self_learning_results_mistral}
  \begin{tabular}{|c|l|r|r|r|}
    \hline
      \textbf{Dataset} & \textbf{Metric} & \textbf{Baseline} & \textbf{LoRA} & \textbf{Dyn-Adapt} \\ \hline
      $\mathbf{Q_H}$ & Hallucination Score & 0.73 & 0.30 & 0.29 \\ \cline{2-5}
      ~ & ROUGE-Lsum & 0.36 & 0.42 & 0.42 \\ \cline{2-5}
      ~ & LLM-Judge & 0.19 & 0.30 & 0.30 \\ \hline
      \textbf{Wiki} & Perplexity & 12.02 & 13.11 & 12.86 \\ \hline
      \textbf{Alpaca} & ROUGE-Lsum & 0.30 & 0.33 & 0.33 \\ \cline{2-5}
      ~ & LLM-Judge & 0.28 & 0.31 & 0.32 \\ \hline
    \end{tabular}
\end{table}

In both approaches, despite the small size of the data and merely three epochs of training, we can observe greatly reduced hallucination on $Q_H$ after self-learning. The model has become more consistent in answering these particular questions. Furthermore, the LLM-Judge score was also increased by 11 percentage points (pp.) in both cases, from 19\% to 30\%. A higher number of training epochs might be needed to make the model's answers more faithful to the ground truth and increase the LLM-Judge score further. Interestingly, the model also gained some improvement on the Alpaca dataset, with 3pp. increase using Plain LoRA and 4pp. increase using Dyn-Adapt.

Both Plain LoRA and Dyn-Adapt exhibited slightly increased perplexity on the Wiki dataset: 1.09 points in the former and 0.84 points in the latter. In both cases, the increase in perplexity was very low, so catastrophic forgetting did not happen. This can be attributed to the adapter technique, which only changed a small number of weights in the model. Plain LoRA experienced a bigger increase in perplexity because the adapter was merged, permanently changing the affected weights. Meanwhile, with Dyn-Adapt, the adapter was enabled only in some examples, while the original weights of the model were unaffected. The significant reduction of hallucination on $Q_H$ after training proves the possibility of self-learning to increase the Known regions and reduce the Unknown regions of the model.

\begin{table}[hbt]
  \centering
  \caption{Phi-3-Mini evaluation results before and after self-learning. "Baseline" shows the metric values before self-learning; "LoRA" shows the metric values after self-learning using Plain LoRA; "Dyn-Adapt" shows the metric values after self-learning using Dynamic-Adapter.}
  \label{table_self_learning_results_phi}
  \begin{tabular}{|c|l|r|r|r|}
    \hline
      \textbf{Dataset} & \textbf{Metric} & \textbf{Baseline} & \textbf{LoRA} & \textbf{Dyn-Adapt} \\ \hline
      $\mathbf{Q_H}$ & Hallucination Score & 0.73 & 0.38 & 0.42 \\ \cline{2-5}
      ~ & ROUGE-Lsum & 0.47 & 0.48 & 0.48 \\ \cline{2-5}
      ~ & LLM-Judge & 0.39 & 0.39 & 0.39 \\ \hline
      \textbf{Wiki} & Perplexity & 8.33 & 11.41 & 11.41 \\ \hline
      \textbf{Alpaca} & ROUGE-Lsum & 0.34 & 0.35 & 0.34 \\ \cline{2-5}
      ~ & LLM-Judge & 0.25 & 0.27 & 0.25 \\ \hline
    \end{tabular}
\end{table}

To ensure the generalizability of our findings, we performed one additional experiment using phi-3-mini as the base of the self-learning LLM, utilizing the corresponding output from the Oracle-Selected experiment. The results are presented in Table~\ref{table_self_learning_results_phi}. Both Plain LoRA and Dyn-Adapt were able to greatly reduce the hallucination on $Q_H$ after self-learning, indicating more consistent answering of those questions. Despite the steep drop in hallucination, the changes in ROUGE-Lsum and LLM-Judge scores were very minuscule; this is because the initial model was already able to answer many questions at least partially correct before self-learning. Notably, the increase in perplexity on the Wiki dataset after self-learning was relatively higher compared to the previous experiment with mistral-instruct. This can be explained by the fact that phi-3-mini is a smaller model; small models tend to be more sensitive to minute changes in the parameters. Still, the perplexity increase is considered quite small. Manual inspection into the model's responses did not reveal any degradation in the text generation quality.

\section{Possible Issues and Potential Extensions}
\label{sec:discuss_issues_extensions}

\textbf{Choosing a knowledge source}. The knowledge source for Knowledge Searching can be a simple API to a search engine or an online wiki. In an organizational environment, it can also be a carefully maintained document database or even a group of human experts tasked with answering the LLM's questions. Finally, the knowledge source can be a stronger LLM, as shown in our experiment, or even several LLMs that are exchanging knowledge with each other, which is discussed in more detail in Section~\ref{sec:discuss_applications}.

\textbf{Dealing with bias, incorrectness, or non-factuality in retrieved data during Knowledge Searching}. Regardless of the knowledge source, a concern during Knowledge Searching is the possibility of biased, incorrect, or non-factual information in the retrieved data. We acknowledge that complete mitigation of these issues is challenging. Still, it can be partially solved by implementing a Curator that is responsible for automatic filtering and scoring of the retrieved data. The Curator would use a classifier model for detecting unwanted types of data and a scorer model for putting more weight on the relevant and preferable data. Alternatively, involving human experts is also an option.

\textbf{Catastrophic forgetting}. 
% Catastrophic forgetting is a risk when performing multiple training cycles on a model. For a finetuned model, the risk is not only related to the loss of language modeling capability but also the loss of alignment and instruction understanding. However, 
Catastrophic forgetting is a risk when performing multiple training cycles in sequence, but in Section~\ref{sec:related_work}, we have pointed out some existing potential solutions. Furthermore, we have experimentally proven that catastrophic forgetting can be avoided by careful training and an effective architectural choice. While the robustness of such solutions still needs to be evaluated for multiple cycles of self-learning, they offer promising starting points. The Dyn-Adapt architecture could be especially effective at preventing catastrophic forgetting in long-term self-learning.

\textbf{Limitations of self-learning}. The goal of self-learning is to enable more efficient model updates, especially in application domains where new information is constantly emerging. This approach excels in dynamic environments. However, it may offer little to no benefits in narrow, closed application domains, such as chatbots that are designed to answer questions about specific products or services. In such cases, the better approach is to simply prepare training data with complete facts of the domain from the beginning. Furthermore, self-learning heavily depends on the quality of the initial model. If the initial model is not sufficiently strong, it may generate nonsensical questions during self-learning. Because of this, it is advised to evaluate the model's Self-Learning Capability before subjecting it to independent learning.

\section{Applications}
\label{sec:discuss_applications}

\textbf{(1) Efficient Training.}
The idea of identification of the Unknown and Known using hallucination score $h(x)$ can be used to filter data used for model training in order to focus on more valuable content.
% - hard to tell which datasets have been used in pretraining/finetuning of a model, Self-Learning can generate questions on which the model absolutely hallucinates

\textbf{(2) Knowledge Exchanging LLMs.} 
Two or more LLMs can exchange their knowledge without external engagement using their self-learning. Model $M_1$ identifies PiUs based on $h_1(x) > \mathit{LIMIT}$. Another model $M_2$ checks $h_2(x) < \mathit{LIMIT}$. If so, $M_2$ provides learning cases related to $x$, which are used by $M_1$ in its self-learning loop. In this way, models exchange only unknown knowledge. Such self-learning with multiple LLMs asking each other would allow efficient knowledge sharing with only useful knowledge.
% - The problem with merging two models' weights (e.g. Model A and Model B) is that: (1) we don't need to merge knowledge that both models already know; (2) incorrectness/uncertainties in A's knowledge, when merged into B, may reduce B confidence in correct knowledge it already has. On the other hand, Self-Learning with multiple LLMs asking each other would allow efficient knowledge sharing where only the necessary knowledge is shared.

\textbf{(3) Direct \emph{Awareness} Optimization.} Model hidden states can be used to detect hallucinations. Then, we can use self-learning to collect examples related to hallucinations and adapt DPO \cite{rafailov2023_dpo} to make the model answer \textit{"I don't know"} instead of hallucinating. Here, the goal is to make the model aware of its own hidden states as the trigger of answering \textit{"I don't know"}, rather than associating concrete concepts/words with \textit{"I don't know"}. This idea is similar to \cite{tian2024_factuality}, 
% https://arxiv.org/abs/2311.08401
though there, the focus was on increasing the model's factuality. In \cite{liang2024_dreamcatcher},
% https://arxiv.org/abs/2401.15449
RLHF was used, even though it highlights a similar idea of making the model aware of its own hidden states. In \cite{liu2023_roe}, they use a reward function to make the model admit \textit{"I don't know"}. Meanwhile, \cite{mckenzie2023_intotheunknown} described a learning task that was coincidentally also termed "Into the Unknown", but their definition is slightly different: it is a learning task where the model is asked to choose from two options the piece of information that would provide new knowledge to the available context, yet the other option might trap the model due to closely resembling the existing knowledge.

Direct Awareness Optimization could improve the self-learning capability of a model. Making the model aware of its own knowledge limit would allow it to ask "better" questions during self-learning; \emph{better} in the sense that the answers to those questions would actually widen the knowledge span of the model. Another aspect of \emph{better} is whether those questions could be considered \emph{meaningful} or not; for example, we might find the question "\emph{How fast does F-16 fly?}" meaningful and could be learned while the question "\emph{How fast do cats fly on Mars?}" not meaningful and does not need to be learned. Such Direct Awareness Optimization is one of our future directions.

\textbf{(4) Learning Multiple Point of Views (PoVs).} 
By adapting DPO and our self-learning concept, it is possible to make the model learn about different PoVs on a certain topic. For example, if topic $T$ has 5 relevant PoVs, the training dataset is then constructed such that a given prompt can have 5 example responses with very similar preferability scores. In another data pair, we alter the prompt by adding some particular context and also increase the preferability score of one example response. This would associate the context of the prompt with a particular PoV.
Similarly, adapting DPO would allow better learning on hard-to-answer open questions. Questions like "\emph{Who can explain the relationship between AI and quantum computing?}" can have multiple valid answers.
% DPO can show this to the model by assigning high preferability scores to the valid answers and low scores to the invalid ones.

\textbf{(5) Decision Making, AGI, Sentience.}
Having a model that automatically learns about the latest trends can be very useful for decision-making systems, for example, for an AI tasked with leading a business or trading.
Self-learning is also a step towards Artificial General Intelligence (AGI). 
% In this way, we can have even more intelligent digital assistants.
Making a model aware of what it knows might lead to a sentient AI.

\section{Limitations}
\textbf{(1) Model's confidence on incorrect knowledge}. We assume that the pretrained models have been subjected to correct knowledge, so consistency of sampled responses would correlate with factuality. This is similar to the assumption in~\cite{tian2024_factuality} and supported by the findings in~\cite{manakul2023_selfcheckgpt}. However, if some incorrect knowledge was repeated in the models' past training data, either accidentally or through deliberate poisoning, the models may become consistent in producing incorrect information. One of our future directions is investigating the integration of a reference-based truthfulness checker, such as FactScore~\cite{min2023_factscore}, in the self-learning loop, which may allow the model to correct wrong understandings and biases by itself.

\textbf{(2) Long-term self-learning}. The focus of this paper is to prove the effectiveness of the methods for the identification of PiUs and the feasibility of self-learning LLM. We have provided a successful demonstration of one full self-learning cycle in our experiment. Still, a deeper study into extensive cycles of self-learning is needed.

\textbf{(3) Experiments were limited to the English language}. We believe that self-learning can be performed in any language. However, further studies would be required to calibrate the brevity coefficient for the SLC score when working with a different language.

% \section{Ethics Statement}
% Self-learning LLM can solve the problem of hallucination and knowledge updating while making efficient use of computing resources, as it trains the model only for knowledge that it does not possess yet. However, some ethical issues may persist, such as bias or incorrectness of newly collected data. Although we have addressed some potential solutions in Section~\ref{sec:discuss_issues_extensions}, we acknowledge that complete mitigation of these issues is challenging due to the fact that latent bias can exist behind the data collected for model training.

\section{Conclusion}
In this work, we show how the concepts of The Known and The Unknown can be utilized to identify atomic pieces of knowledge that an LLM already knows (PiKs) and does not know yet (PiUs). We also propose one extrinsic and three intrinsic methods for the identification of PiUs, which consequently bring up the concept of the self-learning LLM. We formulated the Self-Learning Capability (SLC) Score to gauge the aptitude of an LLM to conduct self-learning.

From the experiments, we concluded that Oracle-Selected is especially effective at enhancing an LLM's capability to Self-Learn. We also found that small models tend to struggle to learn independently. Finetuning or alignment can improve SLC by allowing the model to understand instructions. Yet, if a model's pretraining data contained some instruction examples, the model might be able to Self-Learn even though it has not been explicitly instruction-tuned. Finally, we discussed various possible issues, extensions, and applications of self-learning.

% hidden for triple-blind review
\section*{Acknowledgment}
This work was financed by 
(1) the National Science Centre, Poland, project no. 2021/41/B/ST6/04471;  
(2) the statutory funds of the Department of Artificial Intelligence, Wroclaw University of Science and Technology;
(3) the Polish Ministry of Education and Science within the programme “International Projects Co-Funded”;
(4) CLARIN ERIC – European Research Infrastructure Consortium: Common Language Resources and Technology Infrastructure (period: 2024-2026) funded by the Polish Minister of Science under the programme: "Support for the participation of Polish scientific teams in international research infrastructure projects", agreement number 2024/WK/01;
(5) the European Union under the Horizon Europe, grant no. 101086321 (OMINO). However, the views and opinions expressed are those of the author(s) only and do not necessarily reflect those of the European Union or the European Research Executive Agency. Neither the European Union nor European Research Executive Agency can be held responsible for them.

% \section*{References}

\bibliographystyle{IEEEtran}
\bibliography{main}

\end{document}